\documentclass{article}
\usepackage{spconf,amsmath,graphicx,amssymb,amsfonts}
\usepackage{algorithmicx}
\usepackage[ruled]{algorithm}
\usepackage{algpseudocode}
\usepackage{comment}
\usepackage{xcolor}

\definecolor{newcolor}{rgb}{.8,.349,.1}

\title{Progressive Spatio-Temporal Graph Convolutional Network for Skeleton-Based Human Action Recognition }

\name{Negar Heidari and Alexandros Iosifidis \thanks{This work received funding from the European Union’s Horizon 2020 research and innovation programme under grant agreement No. 871449 (OpenDR). This publication reflects the authors’ views only. The European Commission is not responsible for any use that may be made of the information it contains.}}
\address{Department of Electrical and Computer Engineering, Aarhus University, Denmark}

\begin{document}
\ninept
\maketitle
\begin{abstract}
Graph convolutional networks have been very successful in skeleton-based human action recognition where the sequence of skeletons is modeled as a graph. However, most of the graph convolutional network-based methods in this area train a deep feed-forward network with a fixed topology that leads to high computational complexity and restricts their application in low computation scenarios. In this paper, we propose a method to automatically find a compact and problem-specific topology for spatio-temporal graph convolutional networks in a progressive manner. Experimental results on two widely used datasets for skeleton-based human action recognition indicate that the proposed method has competitive or even better classification performance compared to the state-of-the-art methods while it has much lower computational complexity. 
\end{abstract}

\section{Introduction}
\label{sec:intro}

Skeleton-based human action recognition has attracted great attention in recent years. Compared to other data modalities such as RGB videos, optical flow and depth images, skeletons are invariant to viewpoint and illumination variations, context noise, and body scale while they encode high level features of the body poses and motions in a simple and compact graph structure \cite{han2017space}. Recently, many deep learning approaches have been proposed to model the spatial and temporal evolution of a sequence of skeletons which can be easily obtained by the available pose estimation techniques \cite{shotton2011real,sun2019deep,cao2017realtime}. The existing methods are generally categorized into Convolutional Neural Network (CNN)-based methods, Recurrent Neural Network (RNN)-based methods and Graph Convolutional Network (GCN)-based methods. RNN-based methods mostly utilize Long Short-Term Memory (LSTM) networks \cite{greff2016lstm} to model the temporal evolution of the skeletons which represented as vectors of body joint coordinates \cite{du2015hierarchical,liu2016spatio,shahroudy2016ntu,song2017end,zhang2017view,li2018skeleton}. CNN-based methods represent each skeleton as a 2D matrix which is a suitable input data form for CNNs and model the temporal dynamics by large receptive fields in a deep network \cite{liu2017two,kim2017interpretable,ke2017new,liu2017enhanced,li2017skeleton,li2017skeletonCNN}. The main drawbacks of RNN-based and CNN-based methods are their high model complexity and lack of ability to benefit from the graph structure of skeletons. 
Many GCN-based methods have been proposed in recent years which utilize the well-known pose estimation toolboxes, such as OpenPose \cite{cao2017realtime}, to extract a sequence of skeletons from a video and employ GCN \cite{kipf2016semi} to capture the spatio-temporal connections between the body joints and extract high level features for the human action recognition task. The GCN-based methods have been very successful in processing non-Euclidean data structures and they have reached significant results in skeleton-based human action recognition. 
The first work in this area is Spatio-Temporal Graph Convolutional Network (ST-GCN) \cite{yan2018spatial} which receives a sequence of skeletons as input and after constructing a spatio-temporal graph, it applies several graph convolutions in both spatial and temporal domains to aggregate action-related information in both dimensions. 
2s-AGCN \cite{shi2019two} is proposed on top of ST-GCN and improves the performance in human action recognition by learning the graph structure adaptively using the similarity between the graph joints in addition to their natural connections in a body skeleton. 
Moreover, it uses both joint and bone features of body skeletons in a two-stream network topology that fuses the SoftMax scores at the last layer of each stream to predict the action. AS-GCN \cite{li2019actional} captures richer action-specific correlations by an inference module which has an encoder-decoder structure. DPRL+GCNN \cite{tang2018deep} and TA-GCN \cite{negarTAGCN} aim to make the inference process more efficient by selecting the most informative skeletons from the input sequence. 
GCN-NAS method \cite{peng2020learning} employs the neural architecture search approach to automatically design the GCN model. It explores a search space with multiple dynamic graph modules and different powers of Laplacian matrix (multi-hop modules) in order to improve the representational capacity. However, this method does not optimize the model topology in terms of number of layers and neurons in each layer. Thus, similar to ST-GCN and 2s-AGCN, GCN-NAS also trains a 10-layer model formed by layers of pre-defined sizes and suffers from a long training time and high computational complexity.

Although all the recently proposed GCN-based methods have reached a promising performance in human action recognition, they are not suitable for real-time applications with limited computational resources due to their high computational complexity. 
Finding an optimized and compact topology for the model can make a large step towards addressing the computational and memory efficiency in these methods. However, in order to find the best network topology which is able to reach state-of-the-art performance, an extensive set of experiments training different network topologies is needed. While for other types of deep learning models neural architecture search methods have been proposed for automatically finding good network topologies \cite{iclrZophL17,tran2019heterogeneous}, for ST-GCNs there exist no such methods to date. 
In this paper, we propose a method for automatically finding an optimized and problem-specific network topology for spatio-temporal GCNs (ST-GCNs) which work with non-Euclidean data structures. The proposed method grows the network topology progressively and concurrently optimizes the network parameters. Experimental results show that the proposed method achieves competitive performance compared to the state-of-the-art methods while it builds a compact model with up to $15$ times less number of parameters compared to the existing methods leading to a much more efficient inference process. 

\section{Spatio-temporal GCN}\label{sec:ST-GCN}
\begin{figure}
    \centering
    \includegraphics[width=2.2cm]{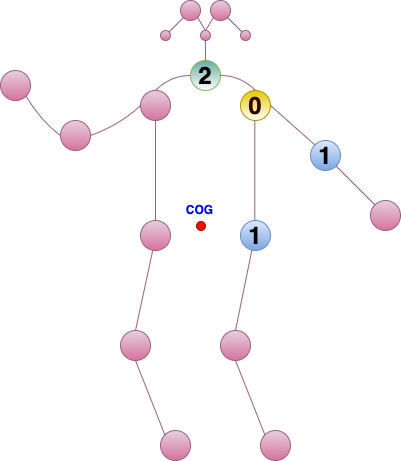}
    \includegraphics[width=3.3cm]{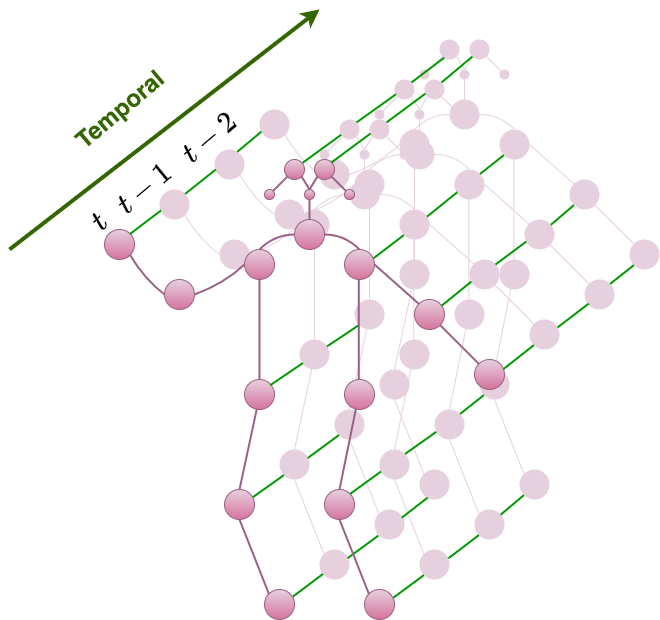}
    \caption{Illustration of Spatio-temporal graph (right), and the neighboring subsets in spatial partitioning process in different colors (left).}
    \label{fig:ST-Graph-Partitioning}
\end{figure}

\vspace{-0.2cm}
In this section, the ST-GCN method \cite{yan2018spatial} is introduced as a baseline of our method. 
It receives a sequence of $T_{in}$ skeletons $\mathbf{X} \in \mathbb{R}^{C_{in} \times T_{in} \times V}$ as input, where $C_{in}$ is the number of input channels, and $V$ is the number of joints in each skeleton. The spatio-temporal graph $\mathcal{G} = (\mathcal{V} ,\mathcal{E})$ is constructed, where $\mathcal{V}$ is the node set of 2D or 3D body joint coordinates and $\mathcal{E}$ encodes spatial and temporal connections between the joints as graph edges. Spatial edges are the natural connections between the body joints and temporal edges connect the same body joints in consecutive skeletons. In such a graph structure, each node can have different number of neighbors. To create a fixed number of weight matrices in convolution operation, the ST-GCN method uses the spatial partitioning process to fix the neighborhood size of each node. This process categorizes the neighboring nodes of each joint into 3 subsets: 1) the root node itself, 2) the neighboring nodes which are closer to the skeleton's center of gravity (COG) than the root node and 3) the nodes which are farther away from the COG than the root node. Figure \ref{fig:ST-Graph-Partitioning} shows a spatio-temporal graph (left) and the $3$ neighboring subsets around a node with different colors. In this setting, each skeleton has $3$ binary adjacency matrices (one for each neighboring node type) which are indexed by $p$ and represent the spatial graph structure in each time step. 

The normalized adjacency matrix $\hat{\mathbf{A}}_p \in \mathbb{R}^{V \times V}$ is defined as: 
\begin{equation}
    \hat{\mathbf{A}}_p = \mathbf{D}^{-\frac{1}{2}}_p\mathbf{A}_p\mathbf{D}^{-\frac{1}{2}}_p,  
\end{equation}
where $\mathbf{D}_{({ii})_p} = \sum_{j}^{V} \mathbf{A}_{({ij})_p} + \varepsilon$, is the degree matrix and $\varepsilon = 0.001$ ensures that there would be no empty rows in $\mathbf{D}_{({ii})_p}$. The adjacency matrix of the root node denoting the nodes' self-connections is set to $\mathbf{A}_1 = \mathbf{I}_V$. In each layer $l$, first a spatial convolution is applied on the output of the previous layer $\mathbf{H}^{(l-1)}$ to update the nodes' features of each skeleton using the layer-wise propagation rule of GCN \cite{kipf2016semi}: 
\begin{equation}
    \mathbf{H}_s^{(l)} = ReLU\left(\sum_{p} (\mathbf{\hat{A}}_p\otimes \mathbf{M}_p^{(l)})\mathbf{H}^{(l-1)}\mathbf{W}_p^{(l)} \right),
    \label{eq:2dConv_s}
\end{equation}
where $\mathbf{M}_p^{(l)} \in \mathbb{R}^{V \times V}$ is a learnable attention matrix which highlights the most important joints in a skeleton for each action and $\otimes$ is the element-wise product of two matrices. The input of the first layer is defined as $\mathbf{H}^{(0)} = \mathbf{X}$. Each neighboring subset is mapped with a different weight matrix $\mathbf{W}_p^{(l)} \in \mathbb{R}^{C^{(l)} \times C^{(l-1)}}$ and $\mathbf{H}_s^{(l)} \in \mathbb{R}^{C^{(l)} \times T^{(l)} \times V}$ with $C^{(l)}$ channels is introduced to the temporal convolution. The temporal dynamics in a sequence of skeletons are captured by propagating the features of the nodes through the time domain. The goal of temporal convolution is to benefit from the motions taking place in each action class to predict the labels. Based on the constructed spatio-temporal graph, each node of a skeleton is connected into its corresponding node in the next and previous skeletons. Therefore, the temporal convolution aggregates node features in $K$ consecutive skeletons by employing a standard 2D convolution with a predefined kernel size $K \times 1$. 

\section{Proposed Method}\label{sec:proposed}\vspace{-0.2cm}
In this section we describe the proposed data-driven method for learning a problem-dependant topology for ST-GCN in an end-to-end manner. The Progressive ST-GCN (PST-GCN) method builds a compact ST-GCN in terms of both width and depth by employing the proposed ST-GCN augmentation module which increases the receptive field in spatial and temporal convolutions progressively. 

\subsection{ST-GCN augmentation module (ST-GCN-AM)}
The goal of each model's layer with temporal convolution, spatial convolution and residual connection is to extract features from a sequence of skeletons by capturing the spatial structure in each skeleton and the motions taking place in each action class.
The proposed ST-GCN-AM is responsible for building a new layer of the ST-GCN model in a progressive manner. 
Let us assume the ST-GCN model is already formed by $l$ layers. ST-GCN-AM receives the existing model and its output $\mathbf{H}^{(l)} \in \mathbb{R}^{C^{(l)} \times T^{(l)} \times V }$ as input and applies an iterative training process (indexed by $t$) by forming a spatial convolution which updates the skeletons' features as follows: 
\begin{equation}
    \mathbf{H}_{s,t}^{(l+1)} = ReLU\left(\sum_{p} (\mathbf{\hat{A}}_p + \mathbf{M}_{p,t}^{(l+1)})\mathbf{H}^{(l)}\mathbf{W}_{p,t}^{(l+1)} \right).
    \label{eq:pro2dConv_s}
\end{equation}
Inspired by 2s-AGCN \cite{shi2019two}, the attention matrix $\mathbf{M}_{p,t}^{(l+1)}$ is added to the normalized adjacency matrix $\hat{\mathbf{A}}_p$ to both highlight the existing connections between the joints and add potentially important connections between the disconnected nodes. 
The spatial convolution defined in Eq. \ref{eq:pro2dConv_s} is a standard 2D convolutional operation which performs $1 \times 1$ convolutions to map the features of the $l^{th}$ layer with $C^{(l)}$ channels, into a new space with $C^{(l+1)}$ channels. At the iteration $t=1$, we set $C^{(l+1)}_t = S$, where $S$ is a hyper-parameter of the method. Therefor, the size of each convolutional filter is $C^{(l)} \times 1 \times 1$ and in order to have $C^{(l+1)}_t$ channels in the output, we need $C^{(l+1)}_t$ different filters of that size. Accordingly, the number of parameters in spatial convolution operation is $C^{(l+1)}_t \times C^{(l)} \times 1 \times 1$. 
The 2D spatial convolution is followed by a batch normalization layer and an element-wise ReLU activation function. The output of the spatial convolution $\mathbf{H}_{s,t}^{(l+1)}$ is introduced to a temporal convolution which aggregates the features through the sequence in order to capture the temporal dynamics. Similar to spatial convolution, it is also a standard 2D convolution with $C^{(l+1)}_t$ different filters, each of size $C^{(l+1)}_t \times K \times 1$, which aggregates the features of $K$ consecutive skeletons with all $C^{(l+1)}_t$ channels and it is followed by a batch normalization layer. The number of skeletons $T^{(l+1)}$ of the layer depends on the stride and kernel size of the temporal convolution. Similar to the ST-GCN \cite{yan2018spatial} and 2s-AGCN \cite{shi2019two} models, a residual connection is added to each layer to stabilize the model by summing up the input of the layer with the layers' output. 
Since the number of channels is changed in each layer by the spatial convolution, the residual connection also transforms the input of the layer by a 2D convolution with $C^{(l+1)}_t$ filters of size $C^{(l)} \times 1 \times 1$ to have the same channel dimension as the layer's output. 
The residual output is added to the temporal convolution output and it is followed by an element-wise activation function ReLU to produce the output of the layer $\mathbf{H}_t^{(l+1)}$ at iteration $t$ which is of size $C^{(l+1)}_t \times T^{(l+1)} \times V$. 
In order to classify the extracted features, the global average pooling (GAP) operation is applied on $\mathbf{H}_t^{(1+1)}$ to produce a feature vector of size $C^{(l+1)}_t \times 1$ which is introduced to the fully connected classification layer. The classification layer is a linear transformation which maps features from $C^{(l+1)}_t$ to $N_{class}$ dimensional subspace where $N_{class}$ denotes the number of classes. 
All the model's parameters are fine-tuned based on back-propagation using the categorical loss value on training data and then, the classification accuracy $\mathcal{A}_t^{(l+1)}$ on validation data is recorded. 

\begin{figure}[!t]
\centering
\centerline{\includegraphics[width=8.5cm]{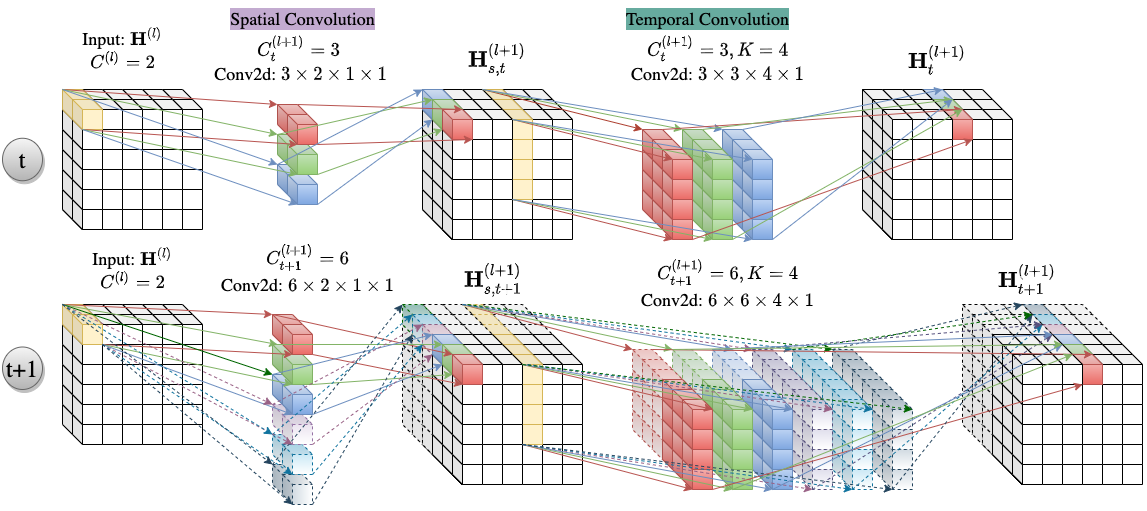}}
\caption{Illustration of the proposed ST-GCN-AM building the model in layer $l$ at iteration $t$ and $t+1$. It is assumed that $C^{(l)} = 2$, $S =3 $ and $K = 4$. Conv2d is the abbreviation for standard 2D convolutional operation.}
\label{fig:Model-Diagram}
\end{figure}

At iteration $t > 1$, the network topology grows in terms of width by increasing the number of output channels in the current layer by $S$ so that the filter size in all the 2D convolutional operations (spatial, temporal and residual) is increased. 
For the spatial convolution and the residual connection, $C^{(l+1)}_t$ convolutional filters of size $C^{(l)} \times 1 \times 1$ are needed. The first $(t-1)S$ filters are initialized by the learned parameters at iteration $t-1$ and the second newly added $S$ filters are initialized randomly. The output shape of the spatial convolution is of size $C^{(l+1)}_t \times C^{(l)} \times 1 \times 1$ which is introduced into the temporal convolution. Therefore, for the temporal convolution $C^{(l+1)}_t$ filters of size $C^{(l+1)}_t \times 1 \times 1$ are needed. As illustrated in Figure \ref{fig:Model-Diagram}, both the width and length of the filters are increased in temporal convolution. The first $(t-1)S$ filters of size $(t-1)S \times 1 \times 1$ are initialized using the learned parameters at iteration $t-1$ (illustrated with solid structure) and the newly added parameters are initialized randomly (illustrated with dotted structure). The size of the spatial attention matrix $\mathbf{M}_{p,t}^{(l+1)}$ does not change, since it only depends on the number of joints in each skeleton. Therefore, all its parameters are initialized to those obtained at iteration $t-1$. The fully connected layer transforms the extracted features with a weight matrix of size $C^{(l+1)}_t \times N_{class}$, in which the first $(t-1)S \times  N_{class}$ parameters are initialized to those from the previous iteration and the remaining ones are initialized randomly. 

After initializing the new weights, all the model parameters are fine-tuned by back-propagation and the new classification accuracy $\mathcal{A}_{t}^{(l+1)}$ of the model on validation data is recorded. Accordingly, the width progression of $(l+1)^{th}$ layer is evaluated based on the performance improvement rate, i.e. $\alpha_{w}= (\mathcal{A}_{t}^{(l+1)}- \mathcal{A}_{t-1}^{(l+1)}) / \mathcal{A}_{t-1}^{(l+1)}$. 
If the model performance has been increased at iteration $t$ by increasing the number of output channels in the $(l+1)^{th}$ layer, its parameters are saved and the next iteration augmenting the layer by adding $S$ new channels starts.  
If increasing the number of output channels in the current layer does not improve the model's performance, i.e. when $\alpha_{w} < \epsilon_{w}$ with $\epsilon_{w} > 0$, the newly added parameters are removed, all the model parameters are set to those obtained at iteration $t-1$, and the width progression for the $(l+1)^{th}$ layer stops. 

\subsection{Progressive ST-GCN method}
The progressive ST-GCN method (PST-GCN) optimizes the network topology in terms of both width and depth by employing the proposed ST-GCN-AM. 
PST-GCN algorithm starts with an empty network and returns the ST-GCN model with an optimized topology. 
The learning process starts by introducing the input data $\mathbf{X} \in \mathbb{R}^{C_{in} \times T_{in} \times V}$ into the ST-GCN-AM which starts building the first layer with a fixed number of output channels and increases the width of the layer progressively until the model's performance converges. In an iterative process, the algorithm builds several layers by employing the proposed module. Let us assume that the width progression is terminated at $(l+1)^{th}$ layer. At this point, the model's performance is recorded and the rate of improvement is given by $\alpha_{d} = (\mathcal{A}^{(l+1)}- \mathcal{A}^{(l)}) / \mathcal{A}^{(l)}$, 
where $\mathcal{A}^{(l)}$, $\mathcal{A}^{(l+1)}$ denote the model's performance before and after adding the $(l+1)^{th}$ layer to the model, respectively. If the addition of the last layer does not improve the model's performance, i.e. when $\alpha_{d} < \epsilon_{d}$ with $\epsilon_{d} > 0$, the newly added layer is removed and the algorithm stops growing the networks' topology. As the final step, the model is fine-tuned using both training and validation sets.

\section{Experiments}\label{sec:experiments}\vspace{-0.2cm}
We conducted experiments on two widely used large-scale datasets for evaluating the skeleton-based action recognition methods. 
The NTU-RGB+D \cite{shahroudy2016ntu} is the largest indoor-captured action recognition dataset which contains different data modalities including the $3$D skeletons captured by Kinect-v2 camera. It contains $56,000$ action clips from $60$ different action classes and each action clip is captured by $3$ cameras with $3$ different views, and provides two different benchmarks, cross-view (CV) and cross-subject (CS). The CV benchmark contains $37,920$ action clips captured from cameras $2$ and $3$ as training set and $18,960$ action clips captured from the first camera as test set. In the CS benchmark, the actors in training and test set are different. The training set contains $40,320$ clips and test set contains $16,560$ clips. In this dataset, the number of joints in each skeleton is $25$ and each sample has a sequence of $300$ skeletons with $3$ different channels each. 
The Kinetics-Skeleton \cite{kay2017kinetics} dataset is a widely used action recognition dataset which contains the skeleton data of $300,000$ video clips of $400$ different actions collected from YouTube. In this dataset each skeleton in a sequence has $18$ joints which are estimated by the OpenPose toolbox \cite{cao2017realtime} and each joint is featured by its $2$D coordinates and confidence score. We use the data splits provided by \cite{yan2018spatial} in which the training and test sets contain $240,000$ and $20,000$ samples, respectively. 

The experiments are conducted on PyTorch deep learning framework \cite{paszke2017automatic} with 4 GRX 1080-ti GPUs, SGD optimizer and cross-entropy loss function. The model is trained for $50$ epochs and batch size of $64$ on NTU-RGB+D dataset, and the learning rate is initiated by $0.1$ and it is divided by 10 at epochs $30$, $40$. The number of training epochs and batch size for Kinetics-Skeleton data set are set to $65$ and $128$, respectively while the learning rate starts from $0.1$ and it is divided by 10 at epochs $45$, $55$. The hyper-parameters $\epsilon_{w}$ and $\epsilon_{d}$ are both set to $1e-4$.
Since we need a validation set to evaluate our progressive method in each step, we divided the training set of each dataset into train and validation subsets. The validation set is formed by randomly selecting $20\%$ of the training samples in each action class and the remaining samples form the training set. 

The performance of the proposed method is compared with the state-of-the-art methods in Tables \ref{table:NTU-ACC} and \ref{table:Kinetics-ACC} on NTU-RGB+D and Kinetics-Skeleton datasets, respectively. Besides, the computational complexity of our method is compared with GCN-based methods in terms of floating point operations (FLOPs) and model parameters in Table \ref{table:Params-FLOPs}. Since the source code of DPRL+GCN is not provided by the authors, it computational complexity is not reported in Table \ref{table:Params-FLOPs}. 
In order to compare our method with state-of-the-art methods which utilize two different data streams such as 2s-AGCN, AS-GCN and GCN-NAS, we also trained our model with two data streams, joints and bones, and reported the results as 2s-PST-GCN. The bone features represent the information of both length and direction of the connection between two joints. 
In two stream training, the PST-GCN finds the optimized network topology for the first data stream (joints) and the SoftMax scores are recorded. The optimized network topology is then fine-tuned using the second data stream (bones) for $30$ epochs and the obtained SoftMax scores obtained in this step are fused by the recorded SoftMax scores to enhance the classification performance. 
The network topology which is used in ST-GCN, 2s-AGCN and GCN-NAS methods, compromise of 10 ST-GCN layers and one classification layer where the number of channels in ST-GCN layers are fixed to $(64,64,64,64,128,128,128,256,256,256)$, respectively. The optimized network topologies which are found by PST-GCN for NTU-RGB+D dataset in CV and CS benchmarks is a 8 layer network. The channel sizes in CV benchmark are $(100,80,100,40,60,80,60,80)$ and in CS benchmarks are $(100,80,60,100,60,100,140,80)$. The network topology which is found for Kinetics-skeleton dataset has 9 layers of sizes $(80,100,100,120,100,120,140,160,180)$, respectively. 

\begin{table}[!t]
	\centering
	\caption{Comparisons of the classification accuracy with state-of-the-art methods on the test set of NTU-RGB+D dataset} \footnotesize
	\label{table:NTU-ACC}
	\resizebox{0.75\linewidth}{!}{
	\begin{tabular}{lccc}
		\hline
		\cline{1-4}
		Method  & CS(\%) & CV(\%) & \#Streams \\
		\hline
		HBRNN \cite{du2015hierarchical} & 59.1 & 64.0 & 5  \\
		Deep LSTM \cite{shahroudy2016ntu} & 60.7 & 67.3 & 1  \\
		ST-LSTM \cite{liu2016spatio} & 69.2 & 77.7 & 1   \\
		STA-LSTM \cite{song2017end} & 73.4 & 81.2 & 1  \\
		VA-LSTM \cite{zhang2017view} & 79.2 & 87.7 & 1   \\
		ARRN-LSTM \cite{li2018skeleton} & 80.7 & 88.8 & 2   \\
		\hline
		Two-Stream 3DCNN \cite{liu2017two} & 66.8 & 72.6 & 2   \\
		TCN \cite{kim2017interpretable} & 74.3 & 83.1 & 1   \\
		Clips+CNN+MTLN \cite{ke2017new} & 79.6 & 84.8 & 1   \\
		Synthesized CNN \cite{liu2017enhanced} & 80.0 & 87.2 & 1   \\
		3scale ResNet152 \cite{li2017skeleton} & 85.0 & 92.3 & 1  \\
		CNN+Motion+Trans \cite{li2017skeletonCNN} & 83.2 & 89.3 & 2  \\
		\hline
		ST-GCN \cite{yan2018spatial} & 81.5 & 88.3 & 1   \\
		DPRL+GCNN \cite{tang2018deep}  & 83.5 & 89.8 & 1  \\ 
		AS-GCN \cite{li2019actional} & 86.8 & 94.2 & 2  \\
		2s-AGCN \cite{shi2019two} & 88.5 & 95.1 & 2  \\
		1s-TA-GCN \cite{negarTAGCN} & 87.97 & 94.2 & 1 \\
		2s-TA-GCN \cite{negarTAGCN} & 88.5 & 95.1 & 2 \\
		GCN-NAS \cite{peng2020learning} & 89.4 & 95.7 & 2  \\
		\hline \hline
		\bf{PST-GCN} & 87.9 & 94.33 & 1  \\ 
		\bf{2s-PST-GCN} & 88.68 & 95.1 & 2  \\ 
		\hline
	\end{tabular}}
\end{table}
\begin{table}[!t]
	\centering
	\caption{Comparisons of the classification accuracy with state-of-the-art methods on the test set of Kinetics-Skeleton dataset}\footnotesize
	\label{table:Kinetics-ACC}
	\resizebox{0.75\linewidth}{!}{
	\begin{tabular}{lccc}
		\hline
		\cline{1-3}
		Method  & Top1(\%) & Top5(\%) & \#Streams\\
		\hline
		Deep LSTM \cite{shahroudy2016ntu} & 16.4 & 35.3 & 1 \\
		TCN \cite{kim2017interpretable} & 20.3 & 40.0 & 1 \\
		ST-GCN \cite{yan2018spatial} & 30.7 & 52.8 & 1 \\
		AS-GCN \cite{li2019actional} & 34.8 & 56.5 & 2 \\
		2s-AGCN \cite{shi2019two} & 36.1 & 58.7 & 2 \\
		1s-TA-GCN \cite{negarTAGCN} & 34.95 & 57.28 & 1 \\
		2s-TA-GCN \cite{negarTAGCN} & 36.1 & 58.72 & 2 \\
		GCN-NAS \cite{peng2020learning} & 37.1 & 60.1 & 2 \\
		\hline \hline
		\bf{PST-GCN} & 34.71 & 57.08 & 1 \\
		\bf{2s-PST-GCN} & 35.53 & 58.2 & 2 \\
		\hline
	\end{tabular}}
\end{table}
\begin{table}[!t]
	\centering
	\caption{Comparisons of the computational complexity with GCN-based state-of-the-art methods. }\footnotesize
	\label{table:Params-FLOPs}
	\resizebox{0.75\linewidth}{!}{
	\begin{tabular}{lccc}
		\hline
		\cline{1-3}
		 Methods & \#\bf{G} \ FLOPs & \#\bf{M} \ Params \\
		\hline
		ST-GCN \cite{yan2018spatial} & $16.7 $ & $3.12 $ \\
		AS-GCN \cite{li2019actional} & $35.92 $ & $7.24 $ \\
		2s-AGCN \cite{shi2019two} & $37.32 $ & $6.9 $ \\
		1s-TA-GCN (NTU-CV-CS) \cite{negarTAGCN} & $5.64$ & $2.24$ \\
		2s-TA-GCN (NTU-CV-CS) \cite{negarTAGCN} & $11.28$ & $4.48$ \\
		1s-TA-GCN (Kinetics) \cite{negarTAGCN} & $9.17$ & $2.24$ \\
		2s-TA-GCN (Kinetics) \cite{negarTAGCN} & $18.34$ & $4.48$ \\
		GCN-NAS \cite{peng2020learning} & $109.26$ & $20.13$ \\ \hline \hline
		\bf{PST-GCN (NTU-CV)} & $7.2 $ & $0.63$ \\
		\bf{2s-PST-GCN (NTU-CV)} & $14.4 $ & $1.26$ \\ \hline
		\bf{PST-GCN (NTU-CS)} & $9.57 $ & $0.92$ \\
		\bf{2s-PST-GCN (NTU-CS)} & $19.14 $ & $1.84 $ \\ \hline
		\bf{PST-GCN (Kinetics)} & $5.67 $ & $1.96 $ \\
		\bf{2s-PST-GCN (Kinetics)} & $11.34 $ & $3.92 $ \\ \hline
	\end{tabular}}
\end{table}

As can be seen in Table \ref{table:NTU-ACC}, the proposed PST-GCN method outperforms all the RNN-based and CNN-based methods with a large margin on NTU-RGB+D dataset in both CV and CS benchmarks. Comprated to GCN-based methods, our method outperforms the ST-GCN and DPRL+GCN methods with a large margin while it has $\times2.3$ less number of FLOPs and and $\times4.95$ less number of parameters than ST-GCN ,according to Table \ref{table:Params-FLOPs}. 
Compared to AS-GCN which trains a two stream model, PST-GCN performs better in both CV and CS benchmarks while it trains the model using only one data stream and it has $\times4.98$ less number of FLOPs and $\times11.49$ less number of parameters. 
The 2s-PST-GCN also outperforms AS-GCN with $\times2.49$ and $\times5.74$ less number of FLOPs and parameters, respectively and it performs on par with 2s-AGCN, 1s-TA-GCN and 2s-TA-GCN in both datasets. The TA-GCN method selects $150$ and $250$ most informative skeletons from the NTU-RGB+D and Kinetics-skeleton datasets, respectively. Therefore, it has less number of FLOPs than our method on NTU-RGB+D dataset. However, our proposed method is still more efficient in terms of number of parameters. 
The best performing method on both datasets is GCN-NAS which has also the highest computational complexity compared to our method and all other state-of-the-art methods. 2s-PST-GCN has competitive performance with GCN-NAS on NTU-RGB+D while it has $\times7.5$ less number of FLOPs and $\times15.97$ less number of parameters. 
On Kinetics-skeleton dataset, Table \ref{table:Kinetics-ACC}, the PST-GCN outperforms the RNN-based methods and ST-GCN method with a large margin while it has less computational complexity. Besides, the 2s-PST-GCN outperforms AS-GCN and performs on par with 2s-AGCN while it has less computational complexity in both cases.

\section{Conclusion}\label{sec:conclusion}\vspace{-0.3cm}
In this paper, we proposed a method which builds a problem-dependant compact network topology for spatio-temporal graph convolutional networks in a progressive manner.
The proposed method grows the network topology in terms of both width and depth by employing a learning process which is guided by the model's performance. Experimental results on two large-scale datasets show that the proposed method performs on par, or even better, compared to more complex state-of-the-art methods in terms of classification performance and computational complexity. Therefore, it can be a strong baseline for optimizing the network topology in ST-GCN based methods.

\vfill
\pagebreak
\bibliographystyle{IEEEbib}
\bibliography{root}

\end{document}